\documentclass{llncs}
\usepackage[section]{algorithm}
\usepackage{algorithmic}
\usepackage{amsmath}
\usepackage{amsfonts}
\usepackage{epsfig}
\usepackage{graphicx}

\begin{document}

\mainmatter

\title{Decomposition Based Search}
\subtitle{A theoretical and experimental evaluation}


\author{W.J. van Hoeve\inst{1} and M. Milano\inst{2}}

\institute{
CWI, P.O. Box 94079, 1090 GB Amsterdam, The Netherlands\\
\email{W.J.van.Hoeve@cwi.nl}\\
\texttt{http://homepages.cwi.nl/\~{ }wjvh/} \and
DEIS, University of Bologna, Viale Risorgimento 2, 40136 Bologna,
Italy\\
\email{mmilano@deis.unibo.it}\\
\texttt{http://www-lia.deis.unibo.it/Staff/MichelaMilano/}
}

\maketitle

\begin{abstract}
In this paper we present and evaluate a search strategy called
Decomposition Based Search (DBS) which is based on two steps: subproblem
generation and subproblem solution. The generation of subproblems is done
through value ranking and domain splitting. Subdomains are explored so as
to generate, according to the heuristic chosen, promising subproblems first.

We show that two well known search strategies, Limited Discrepancy
Search (LDS) and Iterative Broadening (IB), can be seen as special
cases of DBS. First we present a tuning of DBS that visits the same
search nodes as IB, but avoids restarts. 
Then we compare both theoretically and computationally DBS and LDS
using the same heuristic. We prove that DBS has a higher probability
of being successful than LDS on a comparable number of nodes, under
realistic assumptions. 
Experiments on a constraint satisfaction problem and an optimization
problem show that DBS is indeed very effective if compared to LDS.
\end{abstract}

\section{Introduction}

In this work we present a search strategy, called Decomposition Based
Search (DBS) for the solution of constraint satisfaction and optimization
problems. The search strategy is organized in two steps: subproblem
generation and subproblem solution. In the first phase, domain values
are ranked and ordered accordingly for decreasing ranks. Based on this
ranking, domains are partitioned in two or more
subdomains. Subproblems then consist of the initial problem, in
which variables range over one of their subdomains. In the second
phase of DBS the subproblems are being solved. The first subproblem
is considered the most promising one, according to the ranking, to
contain a (good) solution.

The actual generation of subproblems is managed by a tree search in
which we branch on subdomains. Although this tree can be traversed
using any strategy, we prefer to use an LDS-based strategy because it
generates the most promising subproblem first.
If the ranking is accurate, we are likely to find feasible solutions or
good solutions (for optimization problems) early during the search process,
which in the second case is an extremely helpful condition to prove
optimality relatively fast (see also \cite{LodiMilanoRousseauCP03}).
On the other hand, the search process can be stopped once the current
best solution satisfies the user's needs, thus obtaining an incomplete
search strategy.

DBS has several degrees of freedom, whose tuning leads to
different explorations of the subproblem generation tree. Beside the
(traditional) variable and value ordering heuristics, in DBS we have
to tune other parameters concerning the partitioning of domains.
Specifically, the size and the number of subdomains should be tuned
and domains can be partitioned statically or dynamically. Statically
means that domains are divided once for all at the root node while
dynamically means that at each level of the search tree we select a
variable and we partition its domain (which can be already partly
pruned by propagation).

This simple idea was first presented in \cite{focacci_phd}
for scheduling problems modelled through position variables.
This paper can be seen as a generalization and an extension
of our previous work \cite{milano_hoeve02}. In that paper,
we were mainly concerned to show the effectiveness of a reduced cost
based ranking. In this paper, instead, we will theoretically and
computationally evaluate this search strategy and compare it with
other search strategies. To show its behaviour in practice, we apply
DBS to both constraint satisfaction and optimization problems.

Concerning the comparison with other search strategies, we first
present a tuning of DBS such that it traverses the same nodes of the
search tree as Iterative Broadening (IB). Moreover, since DBS avoids
the restarts of IB, it generates less leaf nodes. Next we consider
traditional Limited Discrepancy Search (LDS) and show the equivalence
with DBS when the cardinality of each subdomain is equal to one. In
addition, we show that by considering more than one value in each
subdomain, under realistic conditions, DBS has a higher probability
of being successful than LDS on a comparable number of generated nodes.
Then, we show experimental behaviour of LDS and DBS on the whole
search tree, given a number of probability distributions among the
branches being successful.
Finally, we consider a constraint satisfaction problem, namely the partial
latin square completion problem and a combinatorial optimization problem,
the traveling salesman problem. We apply LDS and DBS to these
problems using the same variable and value ordering heuristics and 
show that DBS outperforms LDS in almost all cases.

The paper is organized as follows.  The next section introduces some 
preliminaries. In Section~\ref{sc:scheme} we propose the subproblem
generation scheme. In Section~\ref{sc:comparison}, we perform a
theoretical comparison with IB and LDS. In Section~\ref{sc:results},
we present a computational study of DBS experimenting it both on
constraint satisfaction and optimization problems. We conclude in
Section~\ref{sc:conclusion}. 

\section{Definitions}
In this work we consider Constraint Satisfaction Problems (or CSPs),
possibly together with an objective function to be optimized.
A Constraint Satisfaction Problem $P$ is defined by the triple
$({\mathcal{X}}, {\mathcal{D}}, {\mathcal{C}})$, where
${\mathcal{X}} = \{x_1, \dots, x_n\}$ is a set of variables,
${\mathcal{D}} = \{D_1, \dots, D_n\}$ is a set of variable domains,
and ${\mathcal{C}} = \{C_1, \dots, C_m\}$ is a set of constraints
$C_j(x_{i_1}, \dots, x_{i_k})$ ($j = 1, \dots, m$) over variables ($k
\leq n$). A {\em solution} $s$ to $P$ is an assignment of each
variable to a value in its domain such that all constraints are
satisfied.

As far as search trees are concerned, we follow the concepts and
vocabulary introduced by Perron \cite{perron99} and Van Hentenryck et
al. \cite{hentenryck_TOCL}. A search tree consists of three disjoint
sets of nodes that are connected to each other: open nodes, closed
nodes and unexplored nodes. The connection between the three is as
follows: 
\begin{itemize}
\item all ancestors of an open node in a search tree are closed nodes, 
\item each unexplored node has exactly one open node as its ancestor,
\item no closed node has an open node as its ancestor.
\end{itemize}
The set of open nodes is called the search {\em frontier}. The search
frontier evolves by so called {\em node expansion}. This operation
removes an open node from the frontier, transforms it to a closed
node, and adds the unexplored children of the node to the frontier. It
corresponds to the {\em branch} operation in the Branch \& Bound
algorithm. In this work, nodes represent CSPs.

\section{Decomposition Based Search}\label{sc:scheme}
Decomposition Based Search is a two-phase search strategy,
consisting of subproblem generation and subproblem solution. In this
section we first give an outline of the strategy. Then details about
subproblem generation are presented. After that, the subproblem
solution is considered.

The input of the DBS algorithm is the problem specification, represented by
a CSP $P_0 = ({\mathcal{X}}, {\mathcal{D}}, {\mathcal{C}})$ and
characteristics of the method that may be defined by the user. Such
characteristics must include a way to evaluate domain values, and a
solution strategy to solve the subproblems. Algorithm~\ref{alg:dbs}
presents the general DBS scheme.
\begin{algorithm}
\caption{Decomposition Based Search}
\label{alg:dbs}
\begin{footnotesize}
\textbf{Input:} CSP $P_0$, variable ordering (used in {\tt choose}),
domain value evaluator {\tt rank}, domain partitioner {\tt partition},
search selector {\tt select}, depth bound $d$,
subproblem solution strategy (used in {\tt solve}), stop criteria {\tt stop}

\begin{algorithmic}
\STATE{open $P_0$}
\WHILE{{\tt stop} not satisfied}
  \STATE{{\tt select} open node $\rightarrow$ $P$}
  \IF{$P$ at depth $d$}
    \STATE{{\tt solve} $P$}
  \ELSE
    \STATE{{\tt choose} variable $x$}
    \STATE{{\tt rank} domain values of $x$}
    \STATE{{\tt partition} domain of $x$}
    \STATE{{\tt expand} $P$}
  \ENDIF
\ENDWHILE
\end{algorithmic}
\end{footnotesize}
\end{algorithm}

\subsection{Subproblem Generation}
The decomposition into subproblems is managed by a search tree. It
can be divided into two parts: the search tree {\em specification}
and the search tree {\em exploration}. The specification defines the
nodes in the search tree, while the exploration defines the way of
traversing those nodes. We will treat both concepts separately.

\subsubsection{Search tree specification}
The subproblem generation tree is specified by variable ordering and 
domain partitioning (based upon some domain value ranking). We list
the basic ingredients for the specification of this search tree.
\vspace{1ex}\\
{\it Variable Ordering - }
This is the traditional variable ordering heuristic which
specifies which variable will be used to expand the current node.
The ordering of variables can be of great
importance during dynamic domain partitioning. One widely used and problem-independent
variable ordering heuristic in CP is the first-fail principle: the
variable with the smallest domain size is selected first. Another
principle is to select the most constrained variable (i.e. the
variable that occurs in the most number of constraints). As usual, problem dependent
heuristics can be used as well.
\vspace{1ex}\\
{\it Domain Value Evaluator - }
In order to rank domain values (function {\tt rank} in Algorithm
\ref{alg:dbs}) we need a domain value evaluator, 
specified by the user.
The ranking is characterized by two levels of accuracy: first, the rank should
give a correct indication on which are the most promising
values. Higher ranks should 
be given to more promising domain values. Second, it should discriminate among values.
Here we introduce the concept of {\em plateaus}: a plateau is a set of values with
the same (or very similar) rank (sometimes also called a {\em tie}).
To perform a theoretical comparison among
search strategies, we assume that the evaluator has a probability distribution
that assigns a certain probability of success to each branch. Plateaus
contain values with the same probability of success.
\vspace{1ex}\\
{\it Domain Partitioner - }
Given a domain value ranking, the user has to specify how the
domain has to be partitioned (function {\tt partition} in Algorithm
\ref{alg:dbs}).
In general, we partition domain $D_i$ into $p$ subdomains $D_i^{(0)},
\dots, D_i^{(p-1)}$ with `best' ranked values in the first subdomain
$D_i^{(0)}$ and the worst ranked values in the last subdomain
$D_i^{(p-1)}$.

The user has to specify the number of subdomains (possibly variable
dependent), and the sizes of the different subdomains. An important point
is that if the heuristics presents plateaus, the domain partitioning
step should in general collect all values in a plateau in the same
subdomain. Thus, the user can partition variable domains according to
equivalent ranks. Another possibility is to partition domains on the
base of their cardinality, e.g., split a domain into two subdomains,
with the 10\% best ranked values in the first and the other 90\% in
the second.
\vspace{1ex}\\
{\it Node Expansion - }
The procedure {\tt expand} $P$ in Algorithm \ref{alg:dbs} generates
$p$ children of a node $P$, being aware of
the selected variable $x_i$ and the domain partition $D_i^{(0)},
\dots, D_i^{(p-1)}$. Based upon the current node $P =
({\mathcal{X}},{\mathcal{D}}, {\mathcal{C}})$, the children $P_r$ are
defined as $P_r = ({\mathcal{X}},{\mathcal{\tilde{D}}}, {\mathcal{C}})$,
where $\tilde{D} = \{D_1, \dots, D_{i-1}, D_i^{(r)}, D_{i+1}, \dots,
D_n \}$, with $r \in \{0, \dots, p-1\}$. The procedure closes node $P$
and opens its children.

\subsubsection{Search tree exploration}
\label{Search-tree-expl} The nodes of the search tree must be visited in a
specific order, based upon the following characteristics.
\vspace{1ex}\\
{\it Search Selector - } The search selector is implemented by
the function {\tt select} that chooses a node to expand from the
frontier. In principle, any search selector ranging from Depth-First
Search to LDS could be used, but LDS fits the most to the idea of
DBS. Instead of LDS, also the Best Bound 
First (BBF) strategy could be suitable to the DBS framework.
BBF is typically applied in a dynamic way,
where it is convenient to recompute the bounds after each node expansion.
When both the ranking heuristic and the bound computation have the
same origin (say reduced costs and a solution from an LP relaxation as
done in \cite{milano_hoeve02}), a dynamic version of LDS will
often behave similar to BBF.

\begin{figure}[t]
\begin{center}
\epsfig{figure=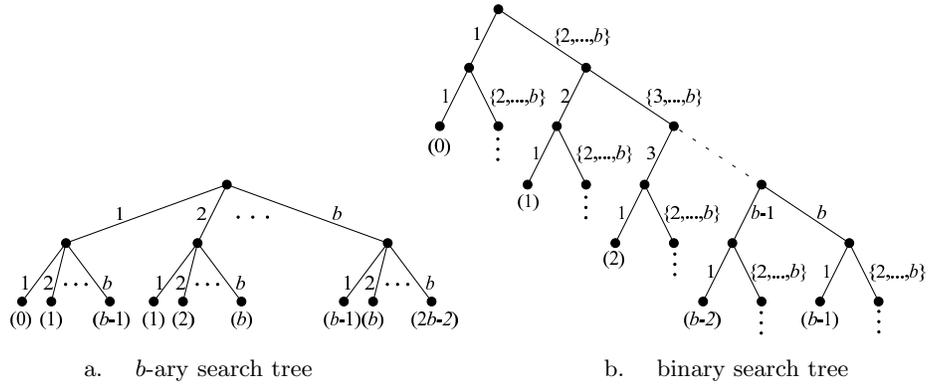,width=\textwidth} \\ \vspace{1ex} a. ~~
$b$-ary search tree \hspace{.3\textwidth} b. ~~ binary search tree
\end{center}
\caption{Two corresponding (partial) search trees. The numbers $1, 2, \dots,
b$ represent domain values. The numbers between parentheses represent the
discrepancy corresponding to the above leaf nodes.} \label{fig:trees}
\end{figure}

In this paper we take into account a LDS Selector. Harvey and Ginsberg
define LDS on binary search trees \cite{harvey95}.
However, we need a general search strategy, therefore we cannot be limited to
binary trees. In the following we recall two version of LDS when
applied to a $b$-ary tree (a search tree with branch width $b$, see
Figure~\ref{fig:trees}.a).

In principle, a $b$-ary search tree  can be mapped onto a binary search tree
(see Figure~\ref{fig:trees}.b), but one has to take into account the depth of
the resulting tree. When $n$ variables ranging on $b$ domain values are
considered, the leftmost path from the root to a leaf in the binary search
tree will be of depth $n$. On the other hand, the rightmost path will be of
depth $n*b$, see Figure~\ref{fig:trees}. This has to be taken into account
when analyzing LDS on a $b$-ary search tree.
For the purpose of this paper, we have chosen not use
binary trees with variable depths, but to maintain a $b$-ary search tree with
fixed depth $n$. For this reason, we need to distribute higher discrepancies
along multiple branches, on multiple depths. A straightforward way to do so
is the following. Each node in the search tree has $b$ branches, ordered by
some heuristic. The branches (ordered
from left to right) contribute a discrepancy of 0 up to $b-1$ and have
a corresponding label weighting the arc, i.e.,  $w_{ij}$ where $i$
represents the depth of the search tree and $j$ the fact that the arc
is ranked $j$. The total discrepancy of a leaf node is
the sum of all branches that form the path from the root node to this
leaf, i.e., $\sum_{i=1}^n w_{ij}$ $j=0..b-1$ (see also Figure~\ref{fig:trees}).

Now we still have a degree of freedom. At each discrepancy $k$ we have two
choices: the first one is to visit nodes labelled with discrepancy $k$
independently from their order. However, when the heuristic used to label
branches represents, for example, preferences, visiting nodes of discrepancy
$k$ in any order is not {\em fair} since we give the same importance to a
choice where $n-1$ variables have the first choice (that suggested by the
heuristics) and one its $k-th$ choice. Therefore, if the heuristics orders
preferences, we have $n-1$ variables completely satisfied and one with its
$k-th$ preference. Among nodes with the same
discrepancy, we could visit first those where the level of un-satisfaction is
more {\em balanced}. Therefore, one could prefer to visit first the node
where $k$ out of 
$n$ variables have a degree of un-satisfaction equal to 1. In this way, we
apply an ordering to nodes with the same discrepancy. Before traversing a
branch whose label is $h$ we have to explore all those paths formed by
branches labelled from 0 up to $h-1$. This preference based ordering
will be used to compare DBS with IB in Section~\ref{ssc:IB}.
\vspace{1ex}\\
{\it Depth Bounding - }
The user may be interested to partition domains only until a certain depth
$d$. At level $d$, the subproblem solution procedure will start.
When $d$ is taken smaller than the depth of the search tree
(together with a LDS search selector and single valued subdomains),
DBS behaves similar to Depth-Bounded Discrepancy Search \cite{walsh_DBDS}.

\subsection{Subproblem Solution}
At a certain level $d$ of depth in the search tree, the user has specified
to solve the subproblem at hand. In order to solve the subproblem,
several methods can be applied, which are often problem
dependent. In the following, we present only a few of the possible
methods to give some insights.

An interesting aspect of the subproblem solution concerns the
use of a different variable and value ordering heuristic with respect
to that used for subproblem generation. Suppose that the first
application of DBS for subproblem generation has grouped domain values
with the same (or similar) ranking value. Using the same heuristic for
the subproblem solution is not very informative  since all values
belonging to the problem have a very similar rank. Thus, it is
convenient to change the heuristic and use for instance one of the
following search strategies.
\vspace{1ex}\\
{\it Standard labelling procedure - }
Solve the subproblem using traditional Depth-First Search. A
motivation for this strategy is that all leaf nodes in the subproblem
are equally likely to be successful with respect to the heuristic
applied to the subproblem generation. Moreover, Depth-First Search
is usually much faster than specialized search strategies.
\vspace{1ex}\\
{\it Iterated application of DBS - }
Another possibility is to apply DBS to the subproblem again.
As stated above, it would be useless to use the same heuristic for
ranking domain values as the one used for subproblem
generation. Instead, one should use a heuristic that captures a
different property of the problem at hand. 
Thus, combining DBS at different levels yields an
effective and simple method for breaking ties.
\vspace{1ex}\\
{\it Local Search - }
An alternative to the use of tree search could be the use of local
moves on a landscape.
In this case, we have to generate some initial solution of the
subproblem, and try to improve it by performing problem dependent
`local moves'. The resulting approach is not `complete'.

\section{Comparison with other approaches} \label{sc:comparison}

This section compares DBS with similar approaches to traverse a search
tree, namely Iterative Broadening (IB) \cite{iterative_broadening} and
Limited Discrepancy Search (LDS) \cite{harvey95}. Note the distinction
between LDS as sole search strategy (single-valued) and LDS as
component of DBS to generate subproblems (multi-valued).

For our comparison we make the following assumptions.
DBS, IB and LDS are applied to the same search tree with fixed branch
width $b$ and depth $n$. Furthermore, we assume that a heuristic
orders the branches in such a way that the probability that branch
$j$ leads to a successful leaf is $p_j$, with $p_1 \geq p_2 \geq \dots
\geq p_b$. As in \cite{harvey95}, this probability is independent of
the depth for the sake of simplicity. Thus the probability that the
first leaf of the ordered search tree is successful is $p_{1}^{n}$.
Finally, a fair comparison is established by the hypothesis that the
heuristic is the same for all three approaches.

As explained in Section \ref{sc:scheme}, we have seen that DBS can be
tuned by fixing the degrees of freedom. In the following we show that
DBS can be tuned in such a way that it is equivalent to IB and even
improves it by avoiding the repeated exploration of some nodes.

Next we consider LDS and show the equivalence with DBS when the
cardinality of each subdomain is taken one. In addition, we show that
by considering more than one value in each subdomain, the first
subproblem generated by DBS has a higher probability of being
successful than LDS under certain conditions.

Finally, we show experimental behaviour of LDS and DBS on the whole
search tree, given a number of probability distributions among the
branches being successful.

\subsection{Iterative Broadening}\label{ssc:IB}
Iterative Broadening (IB) (see \cite{iterative_broadening}) introduces
a breadth cut-off $c_0$ which is the maximum branch-width to explore
in a Depth-First Search (DFS) tree. First $c_0$ is set to some initial
value, and the corresponding search tree is traversed in a DFS
manner. After that, we increase $c_0$ and traverse the extended search tree.
Typically $c_0$ is only increased a small number of times, as to keep the total
nodes to search as low as possible, while still being effective. It is proven
that under certain assumptions IB performs better than DFS
\cite{iterative_broadening}.
One drawback of IB is the redundancy in traversing the search tree. Each time
$c_0$ is increased, the corresponding search tree has to be traversed from
scratch, including the parts that were already visited in previous runs.

The first subproblem generated by DBS can be seen as the first run of
IB, where $c_0$ is the exact number of values to include in the best
subdomain for each variable. If this subtree is being traversed in a
DFS manner, DBS and IB behave equally in this first case. Moreover,
when we apply LDS instead of DFS to traverse the subtree, DBS behaves
provably better than IB (see  \cite{harvey95}).

Suppose that IB increases $c_0$ to $c_1$ up to $c_{\rm max}$. Let
the corresponding consecutive runs be denoted by IB($c_i$). Define
the following domain partitioner (denoted by ($\star$)) for variable
$x_i$, partitioning $D_i$ into $D_i^{(0)}, \dots , D_i^{({\rm max})}$,
where $D_i^{(0)} = \{d_1, \dots, d_{c_0} \}$, and $D_i^{(t)} =
\{d_{c_{t-1} + 1}, \dots d_{c_t} \}$ ($0 < t \leq {\rm max}$).
Here $d_j$ represents the domain value of $D_i$ corresponding to the
$j$-th branch in the ordered search tree. Next, we apply a preference
based ordered (described in Section~\ref{Search-tree-expl}) 
LDS strategy to the subproblem generation tree of DBS. Given a
branch cut-off $c_t$ ($t \in \{0, \dots, c_{\rm max}\}$), let
DBS$(t)^{(\star)}_k$ represent all subproblems 
of discrepancy $k$, using partitioner $(\star)$ restricted to
subdomains $D_i^{(l)}$ with $l \leq t$, in which at least one
$D_i^{(t)}$ is present. Applying this strategy up to discrepancy $k =
t*n$, the subsequent runs of DBS$(t)^{(\star)}_k$ exactly generate
those leaf nodes generated by IB($c_t$) and not generated by
IB($c_{t-1}$), yielding the following theorem. Here
$|$DBS$(t)^{(\star)}_k|$ denotes the number of leaf nodes generated by
DBS$(t)^{(\star)}_k$. Similarly for $|$IB($c_t$)$|$.
\begin{theorem}
Given DBS$(t)^{(\star)}_k$ and IB($c_t$) as described above ($t>0$),
\begin{eqnarray*}
\sum_{k=t}^{t*n} \mbox{\rm prob(DBS$(t)^{(\star)}_k$ successful)} = &
\mbox{\rm prob(IB($c_t$) successful)} - \\
& \mbox{\rm prob(IB($c_{t-1}$) successful)},
\end{eqnarray*} 
and
\begin{displaymath}
\sum_{k=t}^{t*n} |\mbox{\rm DBS}(t)^{(\star)}_k | = 
| \mbox{\rm IB}(c_t) | - | \mbox{\rm IB}(c_{t-1}) |.
\end{displaymath}
\end{theorem}
\begin{proof}
Follows immediately from the above. \qed
\end{proof}
As a consequence of this theorem, note that the redundant
exploration in IB does not appear in the case of DBS.

\subsection{Limited Discrepancy Search}\label{ssc:dbs_lds}
For the definition of Limited Discrepancy Search we refer to
\cite{harvey95} and to our definition for $b$-ary trees given in
Section~\ref{Search-tree-expl}.

When DBS is configured in such a way that the subdomains are
restricted to contain only a single value, traversing the
corresponding tree using a limited discrepancy strategy will be
equivalent to LDS. More interesting is what happens when DBS applies a
limited discrepancy strategy to subdomains of cardinality greater than
one in comparison to LDS on single values. Then one can compare the
total number of generated leaf nodes with the probability of success for
both methods. In the following the {\bf first} subproblem generated by
DBS is compared with LDS on single values.

Assume a domain partitioning of DBS in which the best ranked subdomains
have cardinality $c$ (with $c \leq b$). Let DBS($c$) denote the first
subproblem generated by DBS, corresponding to discrepancy 0. The total
number of leaf nodes of DBS($c$) is $c^n$. The probability of success is
\begin{displaymath}
\mbox{\rm prob(DBS($c$) successful)} = \left(\sum_{i=1}^c p_i \right)^n.
\end{displaymath}

Next the same analysis is performed for LDS. Let LDS($k$) denote the
search subtree consisting of all paths of discrepancy $k$ from the
root to the leaf nodes. At depth $n$ of the search tree, the paths of
discrepancy $k$ can be viewed as partitioning the integer $k$ into
exactly $n$ integers between 0 and $b$. Formally, define the set of
partitions of an integer $k$ as
\begin{displaymath}
\mathcal{P}_{k} = \{ \pi_{i} = (\pi_{i}^{(1)},\pi_{i}^{(2)}, \dots,
\pi_{i}^{(n)}) \; | \; \pi_{i}^{(j)} \; {\rm integer, } \;
0 \leq \pi_{i}^{(j)} \leq b, \; \sum_{j=1}^n \pi_{i}^{(j)} = k \}
\end{displaymath}
Furthermore, each partition $\pi_i$ can occur several times. We denote
its multiplicity as $\mu_i$. The number of leaf nodes of discrepancy
$k$ is thus $\sum_{i=1}^{|\mathcal{P}_k|} \mu_i$. The probability that
these nodes are successful is
\begin{displaymath}
\mbox{\rm prob(LDS($k$) successful)} =
\sum_{i=1}^{|\mathcal{P}_k|} \left( \prod_{j=1}^{n} p_{\pi_{i}^{(j)}}
\right) \mu_i \; .
\end{displaymath}
Let LDS($d$)$_{d=0}^k$ denote the search subtree consisting of all
paths of discrepancy 0 up to $k$ from the root to the leaf nodes. Then
\begin{displaymath}
\mbox{\rm prob(LDS($d$)$_{d=0}^k$ successful)} =
\mbox{\rm $\sum_{d=0}^k$ prob(LDS($d$) successful)}.
\end{displaymath}
Next we present two results that relate the probability of success of
DBS and LDS. The first result considers a search tree in which the
first $c$ branches are ranked equal (called a {\it plateau}). The
second result considers a search tree in which no plateaus occur.
\begin{theorem}\label{th:dbs_lds}
Given $p_1 = p_2 = \dots = p_c > p_{c+1} \geq p_{c+2} \geq \dots \geq p_{b}$,
$1 \leq \tilde{c} \leq c$ and $0 \leq k \leq n(b-1)$:
\begin{displaymath}
\frac{\mbox{\rm prob(DBS($\tilde{c}$) successful)}}{\tilde{c}^n} \geq
\frac{\mbox{\rm prob(LDS($d$)$_{d=0}^k$
    successful)}}{\sum_{d=0}^k\sum_{i=1}^{|\mathcal{P}_d|} \mu_i} .
\end{displaymath}
\end{theorem}
\begin{proof}
The inequality compares the mean probability of success per leaf node
of DBS($\tilde{c}$) and LDS($d$)$_{d=0}^k$. For DBS($\tilde{c}$) this is
$p_1^n$ for all $\tilde{c} \leq c$. This is the same for
LDS($d$)$_{d=0}^k$
when $k < c$. For $k \geq c$, LDS($k$) also uses branches with $p_{c+1}$
up to $p_{b}$ which are strictly smaller than $p_1$ up to $p_c$. Hence
the mean probability of success per leaf node decreases for
LDS($k$). \qed
\end{proof}
\begin{corollary}
Given a problem instance in which for each search variable a plateau
of size $c$ is ranked best, DBS($c$) is more likely to be successful
than LDS($d$)$_{d=0}^k$ on a comparable number of generated leaf
nodes.
\end{corollary}
\begin{proof}
Direct application of Theorem~\ref{th:dbs_lds}. \qed
\end{proof}
We now consider the case when the branches all have strictly
different probabilities of success. In the following theorem we compare
the first subproblem generated by DBS($c$), containing $c^n$ leaf
nodes, with any LDS($d$)$_{d=0}^k$ search tree containing at most
$c^n$ leaf nodes. In other words, we fix $c$ (and hence DBS($c$)), and
make sure that LDS($d$)$_{d=0}^k$ does not generate more leaf nodes than
DBS($c$). In that case, DBS($c$) is more likely to be successful than
LDS($d$)$_{d=0}^k$, assuming $p_1^{n-1}p_{c+1} < p_c^n$.
\begin{theorem}\label{th:strict}
Given $p_1 > p_2 > \dots > p_b$ with $p_1^{n-1}p_{c+1} < p_c^n$,
$n>1$, $1 \leq c \leq b$, $0 \leq k \leq n(b-1)$ and
LDS($d$)$_{d=0}^k$ is allowed to generate at most $c^n$ leaf nodes:
\begin{displaymath}
\mbox{\rm prob(DBS($c$) successful)} \geq
\mbox{\rm prob(LDS($d$)$_{d=0}^k$ successful)},
\end{displaymath}
In particular, equality only holds for the pairs $(c=1,k=0)$,
$(c=b-1,k=n(b-1)-1)$ and $(c=b,k=n(b-1))$.
\end{theorem}
\begin{proof}
It is easily seen that equality holds only for the pairs $(c=1,k=0)$,
$(c=b-1,k=n(b-1)-1)$ and $(c=b,k=n(b-1))$ because for each pair the
generated search trees are equivalent (note that $n(b-1)$ is the
maximum discrepancy of LDS).

The strict inequality comes from the following observation.
LDS($d$)$_{d=0}^k$ can generate at most $c^n$ leaf nodes, and by
nature it differs at least one leaf node from DBS(c) when $n>1$.
Therefore LDS($d$)$_{d=0}^k$ can be built from DBS($c$) by
interchanging DBS($c$) leaf nodes for LDS($d$)$_{d=0}^k$ leaf
nodes. Consider the worst-case interchangement that can occur.
The rightmost leaf node inside the DBS($c$) search tree has the
smallest probability of success within that tree, namely $p_c^n$. The
`first' leaf node of LDS($d$)$_{d=0}^k$ outside the DBS($c$) tree is one
of $n$ in which one branch is of discrepancy $c$ (branch $c+1$) and
the others of discrepancy 0 (branch 1). This leaf node has probability
of success $p_1^{n-1}p_{c+1}$, being the highest outside the DBS($c$)
tree. Since $p_1^{n-1}p_{c+1} < p_c^n$, non-DBS($c$) leaf nodes have a
strictly smaller probability of success than DBS($c$) leaf
nodes. Hence the interchangement will decrease the total probability
of success of LDS($d$)$_{d=0}^k$ with respect to DBS($c$). \qed
\end{proof}

\subsection{Theoretical comparison of LDS and DBS}
\begin{figure}
\begin{center}
\epsfig{figure=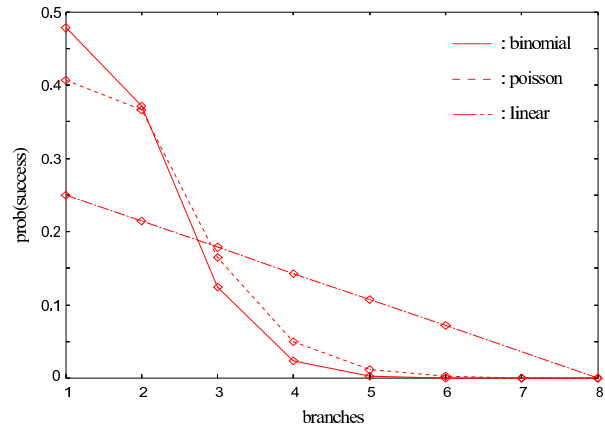,width=.65\textwidth}
\end{center}
\caption{Probability distributions among the branches being successful.} \label{fig:data}
\end{figure}

In this section we compare the behaviour of DBS and LDS on a
whole search tree, given three different probability distributions of
being successful among the branches. We have chosen to compare linear,
poisson and binomial probability distributions, depicted in
Figure~\ref{fig:data}.
This choice is motivated by the different
slopes of the distributions, which will influence the performance of
DBS and LDS. For each probability distribution, also a version
containing plateaus has been used. Such a distribution consists of 4
plateaus of size 2, following the same distribution as its origin
(although being scaled to make the sum among all branches equal to 1).

\begin{figure}[h!]
\begin{center}
\epsfig{figure=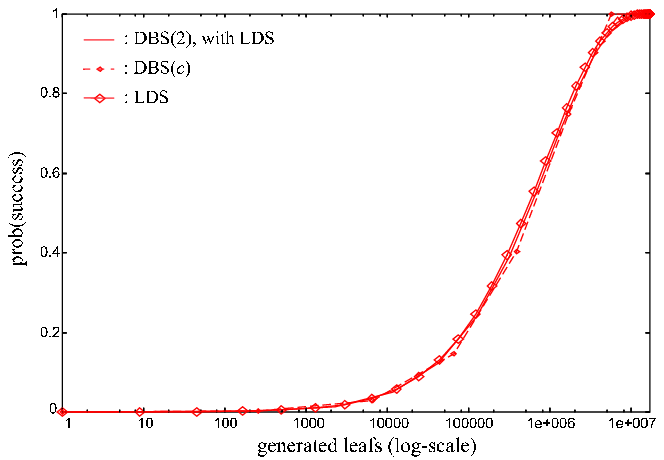,width=.49\textwidth} \hfill
\epsfig{figure=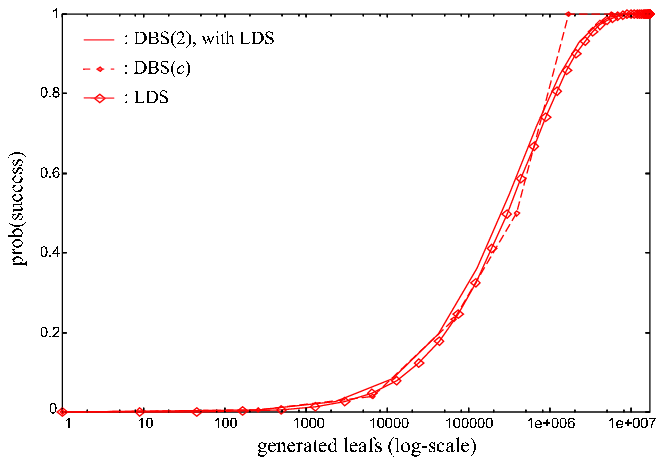,width=.49\textwidth}\\  \vspace{1ex}
a. ~~ linear without plateaus \hspace{17ex}
b. ~~ linear with plateaus \hspace{3ex} \\ \vspace{3ex}
\epsfig{figure=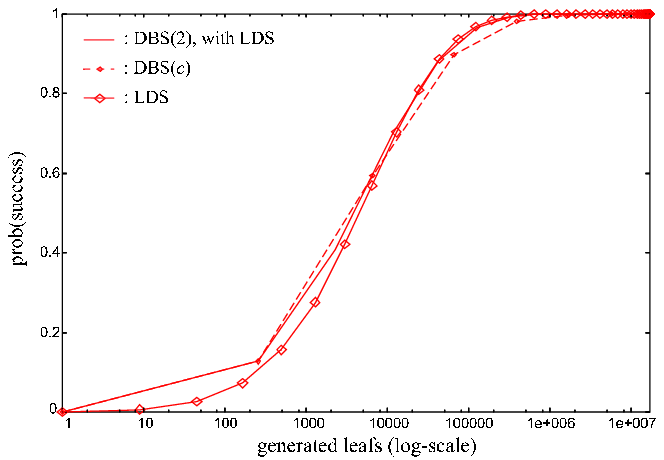,width=.49\textwidth} \hfill
\epsfig{figure=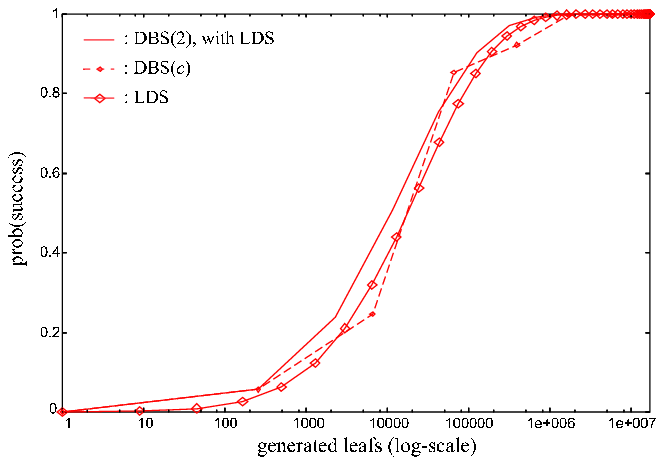,width=.49\textwidth} \\ \vspace{1ex}
c. ~~ poisson without plateaus \hspace{17ex}
d. ~~ poisson with plateaus \hspace{3ex} \\ \vspace{3ex}
\epsfig{figure=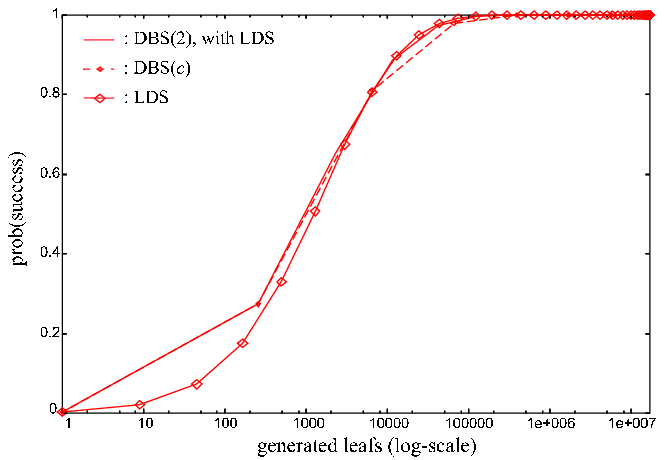,width=.49\textwidth} \hfill
\epsfig{figure=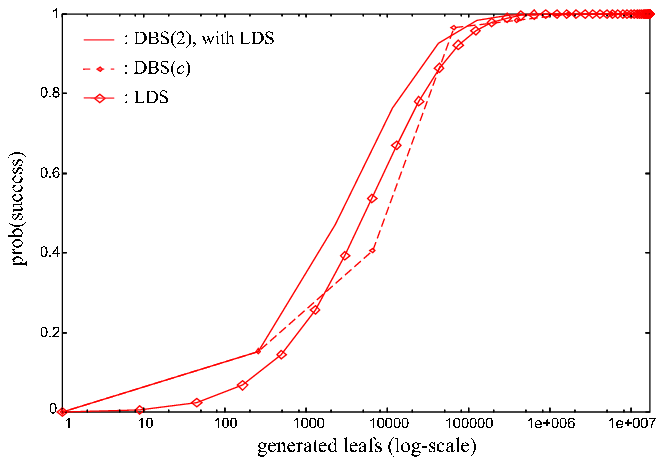,width=.49\textwidth} \\ \vspace{1ex}
e. ~~ binomial without plateaus \hspace{13ex}
f. ~~ binomial with plateaus \hspace{3ex}
\end{center}
\caption{Cumulative probability of success for LDS and DBS on search
  tree of width 8, depth 8 and linear, poisson and binomial
  probability distributions.} \label{fig:cumulative}
\end{figure}

Figure \ref{fig:cumulative} depicts the results of our experiments.
It shows the cumulative probability of success for DBS and LDS. DBS is
plotted in two different ways. The first way represents DBS on
subdomains of size 2 using a limited discrepancy strategy to generate
subproblems (indicated by `DBS(2), with LDS'). The second represents
DBS($c$), the first subproblem generated by DBS, where $c$ is the size
of the best ranked subdomains and ranges from 1 up to $b/2$ (indicated
by `DBS($c$)'). The latter corresponds to DBS($c$) in
Section~\ref{ssc:dbs_lds}.

Figures~\ref{fig:cumulative}.a and \ref{fig:cumulative}.b make use of
a linear descending probability distribution among the branches, with
and without plateaus. One can observe that for this distribution the
performance of DBS and LDS is almost identical, although `DBS(2) with
LDS' performs slightly better than LDS when plateaus are present.

In Figures~\ref{fig:cumulative}.c up to \ref{fig:cumulative}.f  the
used probability distributions are poisson and binomial (both
distributions are scaled such that the sum among the branches equals
1). These figures show a better performance for `DBS(2) with LDS'
compared to LDS, especially in the presence of plateaus. Another
observation concerns DBS($c$) compared to LDS. Even in case of
plateaus, enlarging $c$ does not necessarily make DBS($c$) better than
LDS.

\section{Computational Results} \label{sc:results}
This section presents computational results of two applications for
which we have compared DBS and LDS. The first is the traveling
salesman problem, the second the partial latin square completion
problem. For each application we state the problem, define the applied
heuristic and report the computational results.

For the implementation of the applications we have used ILOG Solver
\cite{ilog51} and Cplex \cite{cplex71} on a Pentium 1Ghz with 256 MB
RAM.

\subsection{Traveling salesman problem}\label{ssc:tsp}
The traveling salesman problem (TSP) is a traditional NP-hard
combinatorial optimization problem. Given a set of cities with
distances (costs) between them, the problem is to find a closed tour
of minimal length visiting each city exactly once.

For the TSP, we have used a heuristic similar to the one used
in~\cite{milano_hoeve02}. It relies on the reduced cost matrix that
originates from the solution of a linear relaxation (an assignment
problem) inferred from the TSP.
The heuristic ranks those values best that are associated to the
lowest reduced costs. Intuitively this is motivated by the fact
that those values will contribute the least to an optimal
tour. Differently from \cite{milano_hoeve02} we now apply this
heuristic in a dynamic way. This means that the subdomains are not
selected beforehand (statically) but during the subproblem
generation. This approach avoids the inclusion of already pruned
domain values. Secondly, for each variable we only include values that
have the same (lowest) reduced cost, instead of a range of low reduced
costs.

The further tuning of DBS consists of the following. The subproblems
are generated using a limited discrepancy strategy, without preference
ordering concerning the discrepancies. Subproblems are being solved
using depth-first search, since all leaf nodes can be considered to
have equal probability of success.

To compare LDS and DBS fairly, we stop the search as soon as an
optimal solution has been found. The proof of optimality should not be
taken into account, because it is not directly related to the
probability of a branch being successful.

\begin{table}[t]
\begin{center}
\begin{footnotesize}
\begin{tabular}{|p{1.5cm}|p{1.3cm}p{1.3cm}p{.8cm}|p{1.3cm}p{1.3cm}p{.8cm}|}
  \hline \hline
 & \multicolumn{3}{|c|}{\bf LDS} & \multicolumn{3}{|c|}{\bf DBS} \\
instance & time (s) & fails & discr & time (s) & fails & discr  \\ \hline
gr17 & 0.08 & 36 & 2 & 0.02 & 3 & 0 \\
gr21 & 0.16 & 52 & 3 & 0.01 & 1 & 0 \\
gr24 & 0.49 & 330 & 5 & 0.01 & 4 & 0 \\
fri26 & 0.16 & 82 & 2 & 0.01 & 0 & 0 \\
bayg29 & 8.06 & 4412 & 8 & 0.07 & 82 & 1 \\
bays29 & 2.31 & 1274 & 5 & 0.07 & 43 & 1 \\
dantzig42 & 0.98 & 485 & 1 & 0.79 & 1317 & 1 \\
swiss42 & 6.51 & 2028 & 4 & 0.08 & 15 & 0 \\
hk48 & 190.96 & 35971 & 11 & 0.23 & 175 & 1 \\
brazil58 & N.A. & N.A. & N.A. & 0.72 & 770 & 1 \\ \hline \hline
\end{tabular}\\ \vspace{1ex}
N.A. means `not applicable' due to time limit (900 s).
\end{footnotesize}
\end{center}
\caption{Results for finding optima of TSP instances (not proving
  optimality).}\label{tb:tsp}
\end{table}
The results of our comparison are presented in Table~\ref{tb:tsp}.
The instances are taken from TSPLIB \cite{TSPLIB} and represent
symmetric TSPs. For LDS and DBS, the table shows the time and the
number of fails (backtracks) needed to find an optimum. For LDS, the
discrepancy of the leaf node that represents the optimum is given. The
discrepancy of the subproblem that contains the optimum is reported
for DBS.

For all instances but one, DBS performs much better
than LDS. Both the number of fails and the computation time are
substantially less for DBS. Observe that for the instance {\tt
  dantzig42} LDS needs less fails than DBS, but uses more time. Here
is where the depth-first search strategy for solving the DBS
subproblems pays off. It can visit almost three times more nodes in
less time, because it lacks the LDS overhead.

\subsection{Partial latin square completion problem}
The partial latin square completion problem (PLSCP) is a well known
NP-complete combinatorial satisfaction problem. A latin square is an
$n \times n$ square in which each row and each column is a permutation
of the numbers $\{1, \dots, n\}$. For example:
\begin{center}
\begin{tabular}{|ccccccccc|} \hline
& 2 & & 4 & & 3 & & 1 & \\
& 1 & & 3 & & 2 & & 4 & \\
& 4 & & 2 & & 1 & & 3 & \\
& 3 & & 1 & & 4 & & 2 & \\ \hline
\end{tabular}\\
\end{center}
is a latin $4 \times 4$ square. A partial latin square is a partially
pre-assigned square. The PLSCP is the problem of extending a partial
latin square to a feasible (completely filled) latin square.

As heuristic we have used a simple first-fail principle for the
values, i.e. values that are most constrained are to be considered
first. This means, a value that occurs the most inside a partial
latin square is ranked best. Hence the rank of a value is taken equal
to the number of the value's occurrences in the partial latin square,
and a higher rank is regarded better.

In our implementation, DBS groups together values of the same rank to
generate subproblems, using a limited discrepancy strategy without
preference ordering concerning the discrepancy. The subproblems are
being solved using a depth-first strategy, since we consider all
values of the same rank to be equally successful. Furthermore, the CSP
that models the PLSCP uses {\tt alldifferent} constraints on the rows
and the columns, with maximal propagation. The maximal {\tt
  alldifferent} propagation (achieving hyper-arc consistency) is of
great importance for solving the PLSCP as a CSP. With less powerful
propagation, the considered instances are practically unsolvable.

\begin{table}[t]
\begin{center}
\begin{scriptsize}
\begin{tabular}{|p{3.3cm}|p{1.3cm}p{1.1cm}p{.8cm}|p{1.1cm}p{1.1cm}p{.8cm}|}
  \hline \hline
 & \multicolumn{3}{|c|}{\bf LDS} & \multicolumn{3}{|c|}{\bf DBS} \\
instance & time (s) & fails & discr & time (s) & fails & discr  \\ \hline
{\tt bpls.order25.holes238} & 2.36 & 668 & 5 & 1.09 & 746 & 5 \\
{\tt bpls.order25.holes239} & 0.49 & 15 & 1 & 0.42 & 2 & 1 \\
{\tt bpls.order25.holes240} & 1.17 & 179 & 4 & 0.86 & 893 & 4 \\
{\tt bpls.order25.holes241} & 3.31 & 772 & 3 & 4.70 & 3123 & 4 \\
{\tt bpls.order25.holes242} & 2.41 & 537 & 3 & 1.80 & 1753 & 4 \\
{\tt bpls.order25.holes243} & 4.06 & 1082 & 4 & 3.96 & 2542 & 4 \\
{\tt bpls.order25.holes244} & 1.33 & 214 & 3 & 2.99 & 2072 & 4 \\
{\tt bpls.order25.holes245} & 9.40 & 2308 & 6 & 10.66 & 12906 & 7 \\
{\tt bpls.order25.holes246} & 2.01 & 401 & 5 & 2.22 & 1029 & 4 \\
{\tt bpls.order25.holes247} & 258.91 & 69105 & 6 & 11.66 & 5727 & 4 \\
{\tt bpls.order25.holes248} & 33.65 & 6969 & 5 & 0.68 & 125 & 2 \\
{\tt bpls.order25.holes249} & 212.76 & 60543 & 11 & 101.46 & 85533 & 8 \\
{\tt bpls.order25.holes250} & 2.45 & 338 & 2 & 0.83 & 687 & 3 \\
{\tt pls.order30.holes328} & 273.53 & 32538 & 4 & 82 & 14102 & 3 \\
{\tt pls.order30.holes330} & 21.79 & 2756 & 3 & 25.15 & 5019 & 3 \\
{\tt pls.order30.holes332} & 235.40 & 30033 & 5 & 56.94 & 9609 & 3 \\
{\tt pls.order30.holes334} & 4.18 & 256 & 2 & 6.09 & 843 & 2 \\
{\tt pls.order30.holes336} & 1.73 & 69 & 2 & 0.76 & 12 & 1 \\
{\tt pls.order30.holes338} & 49.17 & 5069 & 3 & 29.41 & 8026 & 3 \\
{\tt pls.order30.holes340} & 1.68 & 91 & 2 & 0.81 & 66 & 2 \\
{\tt pls.order30.holes342} & 28.40 & 3152 & 3 & 5.41 & 600 & 2 \\
{\tt pls.order30.holes344} & 9.05 & 605 & 2 & 8.35 & 1103 & 2 \\
{\tt pls.order30.holes346} & 2.15 & 101 & 2 & 3.76 & 482 & 2 \\
{\tt pls.order30.holes348} & 43.80 & 2658 & 2 & 32.86 & 2729 & 2 \\
{\tt pls.order30.holes350} & 1.16 & 46 & 1 & 0.80 & 12 & 1 \\
{\tt pls.order30.holes352} & 5.10 & 288 & 2 & 0.95 & 32 & 1 \\ \hline
sum & 1211.45 & 220793 & 91 & 396.62 & 159773 & 81 \\
mean & 46.59 & 8492.04 & 3.50 & 15.25 & 6145.12 & 3.12 \\ \hline \hline
\end{tabular}
\end{scriptsize}
\end{center}
\caption{Results for PLS completion problems.}\label{tb:latin}
\end{table}
In Table~\ref{tb:latin} we report the performance of LDS and DBS on a
set of partial latin square completion problems. The instances are
generated with the PLS-generator {\tt lsencode} by Gomes et
al. \cite{lsencode}. Following remarks made in \cite{gomes_color02},
our generated instances are such that they are `difficult' to
solve. The instances {\tt bpls.order25.holes}$m$ are balanced $25
\times 25$ partial latin squares, with $m$ unfilled entries (around
38\%). Instances {\tt pls.order30.holes}$m$ are unbalanced $30 \times
30$ partial latin squares, with $m$ unfilled entries (around 38\%).

Again, we report the time and the number of fails (backtracks) needed
to find a solution for both LDS and DBS. The discrepancy of the leaf
node that represents the solution is reported for LDS, for DBS this is
the discrepancy of the subproblem that contained the solution.
Although DBS performs much better than LDS on average, the results are
not homogeneous. For some instances LDS even found a solution at a
lower discrepancy level than DBS. This can be explained by the pruning
power of the {\tt alldifferent} constraint. Because DBS branches on
subdomains of cardinality larger than one, the {\tt alldifferent}
constraint will remove less inconsistent values compared to branching
on single values, as is the case with LDS. Using DBS, such values
will only be removed inside the subproblems.

As was already mentioned in Section~\ref{ssc:tsp}, DBS effectively
exploits the depth-first strategy which it is allowed to use to solve
the subproblems. For a number of instances, DBS finds a solution
earlier than LDS, although making use a higher number of fails.

\section{Conclusion}\label{sc:conclusion}
In this paper, we presented a theoretical and experimental 
evaluation of an effective search strategy, Decomposition Based Search
(DBS), based on value ranking and domain partitioning. We have shown
that DBS can be tuned to implement two well known search strategies, namely 
Iterative Broadening and Limited Discrepancy  Search. 
Concerning IB, we show that DBS explores the same number of nodes of
each IB iteration, but avoids restarts. As far as LDS is concerned, we
prove that DBS has a higher probability of success on a comparable
number of nodes. Experimental result on the partial latin square
completion problem and on the traveling salesman problem show that DBS
outperforms LDS in almost all cases.

\end{document}